%% file: main.tex
\documentclass{article} 
\usepackage{iclr2027_conference,times}

\input{math_commands.tex}

\usepackage{hyperref}

\usepackage{url}
\usepackage{graphicx}

\usepackage[super]{nth}
\usepackage{amsmath}
\usepackage{amssymb}
\usepackage{mathtools} 
\usepackage{amsthm}
\usepackage{mathtools} 
\usepackage{caption}
\usepackage{physics}
\usepackage{subfigure}
\usepackage{makecell}
\usepackage{arydshln}
\usepackage{bm}
\usepackage{pifont}
\usepackage{algorithm}
\usepackage{algpseudocode}
\usepackage{booktabs}
\usepackage{wrapfig}

\usepackage{graphicx}
\usepackage{subcaption}
\usepackage{multirow}

\title{Subspace Node Pruning: Retraining-Free Neural Network Compression via Orthogonal Subspace Projections}

\author{Joshua Offergeld, John Chidiac, Marcel van Gerven, Nasir Ahmad \\
Department of Machine Learning and Neural Computing\\
Donders Institute for Brain, Cognition and Behaviour\\
Radboud University, Nijmegen, the Netherlands\\
\texttt{\{joshua.offergeld, nasir.ahmad\}@ru.nl}
}

\iclrfinalcopy 

\begin{document}

\maketitle

\begin{abstract}
Node pruning is the art of removing computational units such as neurons, filters, attention heads, or even entire layers to significantly reduce inference time while retaining network performance.
In this work, we show that a broad family of importance measures, from activation-based to Fisher-based, share a common quadratic form.
Through a lower-triangular change of basis, this form yields a pruning subspace in which entire nodes can be removed while the impact of the lost units is simultaneously recovered via linear least squares.
Furthermore, the importance measures within our subspace are decoupled and additive through an orthogonalization of the layer activations, allowing multi-node pruning within layers without interference between importance scores.
Our method, Subspace Node Pruning (SNP), outperforms the compared baselines on VGG-16 and ResNet-50 on ImageNet. On OPT-125M and LLaMA-2 7B, SNP matches the state-of-the-art LLM Surgeon at a 40-shot schedule and outperforms it whenever fewer shots are used, all without any retraining after pruning.
Code is available at \href{https://github.com/EfficientNeuralSystems/subspace-node-pruning}{https://github.com/EfficientNeuralSystems/subspace-node-pruning}.
\end{abstract}

\section{Introduction}

Alongside hardware optimizations~\citep{jouppi2018motivation, choquette2021nvidia}, software frameworks~\citep{tensorflow2015-whitepaper, jax2018github, paszke2019pytorch}, and quantization~\citep{krishnamoorthi2018quantizing, gholami2022survey}, exist methods which directly modify and compress neural network models to reduce computational cost while maintaining accuracy.

Neural network pruning is a particular approach which aims to remove parameters from a trained model in order to reduce the computational execution (inference) time while maintaining its performance.
Unfortunately, unstructured approaches to pruning can result in models with arbitrarily sparsified weight matrices whose multiplication cannot easily be accelerated at compute time.
It is therefore desirable to prune structured aspects of neural networks models: whole neurons, convolutional filters, transformer heads, or other structured groups of parameters.
Herein, we refer to any of these sub-parts of networks as network `units' or `nodes'.
Two broad approaches have emerged for network pruning.
Pruning is either applied to pre-trained networks while trying to maintain their performance, or is carried out as an iterative process during training.
We limit our investigation to the pruning of pre-trained networks, assuming that one starts out with a well-performing model and does not wish to carry out any further (re)training.

\textbf{Importance Scores}\;\;When choosing which units of a network to prune, there must first be an attribution of the relative `importance' of each network unit.
A number of methods use the magnitude of weights as a proxy for an importance score, assuming that smaller weights have a smaller impact on a network's computation~\citep{li2016pruning}.
Methods of greater complexity are data-driven: they use training data for forward passes, and in some cases backward passes, to refine their importance measures.
The simplest of these methods score units by their activations alone~\citep{hu2016network, berisha2025variance}, or by their activations combined with their outgoing weights~\citep{he2017channel, luo2017thinet, sun2023simple}.
Loss-based measures instead use a first-order Taylor expansion of the loss~\citep{molchanov2016pruning, molchanov2019importance}, Fisher information~\citep{theis2018faster}, or even downstream network importances~\citep{yu2018nisp}.
Second-order measures date back to Optimal Brain Damage and Optimal Brain Surgeon~\citep{lecun1989optimal, hassibi1992second}, and have been made tractable for large networks by Kronecker-factored curvature approximations~\citep{martens2015optimizing, wang2019eigendamage}, more recently for structured pruning of large language models in LLM Surgeon~\citep{van2024llm}.

Other approaches look at feature maps to determine filters which have greatest task-relevant information~\citep{liu2023filter, zhang2022fchp} or various correlation measures which treat high correlation between units as a sign of redundancy~\citep{mariet2015diversity, ayinde2019redundant, cuadros2020filter, kim2020neuron, goldberg2022pruning,  zhang2022fchp, he2019filter}.

\textbf{Reconstruction}\;\;After having pruned a network, it is often possible to analytically determine specific changes which can be made to the remaining parameters in a network to account for the removed unit.
We refer to this process as `reconstruction'.
Unlike retraining or finetuning, reconstruction is a single step modification to a network.
Optimal Brain Surgeon~\citep{hassibi1992second} derives such a change in closed form from the curvature of the loss, an approach that LLM Surgeon~\citep{van2024llm} scales to structured pruning of large language models.
When several units are removed at once, however, their costs and compensations interact.
Optimal Brain Compression~\citep{frantar2022optimal} and ZipLM~\citep{kurtic2023ziplm} therefore remove units greedily, one at a time, updating the inverse curvature after each removal so that later costs account for earlier ones.

Choosing which nodes to prune and which can be best reconstructed is also an important consideration.
\citet{mariet2015diversity} identified redundant nodes by using determinantal point processes, pruned these nodes, and recovered their impact on a network by linear least squares (LLS).
Similarly,~\citet{he2017channel} selected units based on LASSO regression and again used LLS to approximate pruned units.
\citet{chee2022model} instead select the units to keep by pivoted QR, an interpolative decomposition, and reconstruct the remaining units from them by least squares.
Other work uses less expressive approaches, with single scalar values used for reconstruction per node~\citep{luo2017thinet}, or even attempt data-free approaches to reconstructing unit activity~\citep{theus2024towards}. 
Other work goes further still, into a non-linear least squares solution attempt using evolutionary algorithms~\citep{chin2018layer}, or application of approximate reconstruction even to large language models~\citep{frantar2023sparsegpt, li2024greedy}, for instance by compensating removed channels through bias terms~\citep{an2024fluctuation}.

\textbf{Transformer Pruning}\;\;Methods for pruning transformers go beyond estimating local or global importance, often pruning nodes at several granularities simultaneously.
\citet{zheng2022savit} designed an optimization problem to prune attention heads, embedding dimensions and the hidden layers of the feedforward networks.
They combined a second-order Taylor estimation with evolutionary algorithms to optimize a pruning mask.
\citet{yang2023global} went even further by pruning \textit{query} and \textit{key} embedding dimensions independent of the \textit{value} embedding dimension.
Several other recent influential methods train networks for estimating global sparsity~\citep{yu2023x, yang2023pruning, yu2022unified} and therefore fall outside of the scope of our setting. 
Dynamic removal of tokens/patches to reduce the inference time for a given sample is also an area under increasing investigation~\citep{rao2021dynamicvit, song2022cp, wang2022vtc, liang2022not} but this prunes inputs rather than the network directly.

\textbf{This work}\;\;
Herein we describe and empirically validate a novel framework for the entire pipeline of node importance scoring, pruning, and reconstruction.
Our core contributions are:
\begin{enumerate}
\item We provide a common form for a range of importance measures. Activation-based (zeroth-order), first-order, and second-order importance measures share a common quadratic form that differs only in a layer-wise sensitivity. The importance measure is therefore an interchangeable component, and the properties below hold for all of them.
\item We show how the orthogonalization of unit activations makes importance scores decoupled and additive. By using a lower-triangular orthogonalization transform, each unit's score only reflects the importance not already captured by the units preceding it. Combined with a unit ordering method, multiple units can then be pruned within a layer while the total `lost importance' is exactly the sum of their individual importances.
\item We construct a subspace in which pruning is carried out linearly reconstructs lost activity, and is equivalent to pruning network units. This differentiates our subspace from other subspace methods (like PCA, ZCA etc) in that pruning in our subspace reduced network layer widths and provides measurable efficiency gains.
\end{enumerate}
Existing methods either score units independently, neglecting correlations between importance scores~\citep{van2024llm}, or account for these correlations by removing units sequentially~\citep{frantar2022optimal, kurtic2023ziplm}, which requires a separate update after every removal.
In comparison, our subspace has exactly additive pruning costs allowing multi-unit pruning while taking into account correlations, and is efficient in not requiring a re-computation of the subspace if the number of units being pruned is changed.
We demonstrate the effectiveness and computational efficiency of our approach in comparison to the state of the art by application to ImageNet-trained VGG-16 and ResNet-50, as well as OPT and LLaMA-2 language models evaluated on the WikiText dataset, all without any retraining.

\section{Methods} 
Consider a typical deep neural network (DNN) architecture, in which the outputs at each layer, $l \in \{1,\ldots,L\}$, are defined as
$
\vb{X}_l = f_l(\vb{Y}_l) = f_l(\vb{W}_{l-1} \vb{X}_{l-1})
$, 
where $\vb{X}_l \in \mathbb{R}^{N_l \times K}$ is a matrix of outputs for layer $l$ consisting of $N_l$ units for $K$ samples.
These layer outputs are composed based on a matrix multiplication of the previous layer outputs, from weights $\vb{W}_l \in \mathbb{R}^{N_{l+1} \times N_{l}}$, and by an element-wise transfer function $f_l(\cdot)$.
Let $\mathcal{L}$ denote the total loss of the network over the $K$ samples.
We consider fully-connected deep neural networks to introduce our approach, however, we also apply our theory to convolutional and transformer architectures in the results that follow.

\subsection{Factorizing neural contributions}
\label{sec:factorising}
Importance scores, whether based upon node activation or loss impact via first/second-order Taylor approximation, are approximate (or heuristic) measures of the degree of impact that removing an individual unit has on a network's performance.
A broad set of importance scores can be expressed in the form that follows, letting $\vb{x}_{l}^{(k)}$ be the activity of all units at layer $l$ for sample $k$, we define the importance of all units in a layer as
\begin{equation*}
 I_l = \sum_{k=1}^{K} \left(\vb{A}_l^{(k)} \vb{x}_l^{(k)}\right)^\top \left(\vb{A}_l^{(k)} \vb{x}_l^{(k)}\right),
\end{equation*}
where $\vb{A}_l^{(k)}$ is the sample-wise sensitivity matrix that maps the unit activities onto the space in which their impact is measured for importance.
The choice of $\vb{A}_l^{(k)}$ recovers commonly used importance measures as shown in Table~\ref{table:sensitivities}.

\begin{table}[t]
  \centering
  \small
  \caption{Common importance measures expressed in terms of the activities
  at layer $l$, transformed by a sample-wise sensitivity matrix
  $\vb{A}_l^{(k)}$. The per-unit importance
  $I_{l,i}$ is the contribution of unit $i$ to $I_l$, excluding cross terms
  between units. Here
  $\vb{g}_l^{(k)} = \partial \mathcal{L} / \partial \vb{x}_l^{(k)}$ is the gradient of
  the loss with respect to the unit activities, and $\vb{H}_l^{(k)}$ is a curvature estimate \citep{martens2020new} of the loss with respect to the layer's output $\vb{y}_{l+1}^{(k)}$: the Gauss--Newton Hessian, the true Fisher, the empirical Fisher $\vb{\delta}_l^{(k)}\vb{\delta}_l^{(k)\top}$ with $\vb{\delta}_l^{(k)} = \partial\mathcal{L}/\partial\vb{y}_{l+1}^{(k)}$, or its sample average $\bar{\vb{H}}_l$ as in K-FAC~\citep{martens2015optimizing}. Since $\vb{g}_l^{(k)} = \vb{W}_l^\top\vb{\delta}_l^{(k)}$, the empirical
  Fisher recovers the first-order measure up to a factor of $\flatfrac12$.
  The per-unit importance $I_{l,i}$ is the contribution of unit $i$ to the
  layer-wide importance $I_l$, excluding cross terms between units, where
  $x_{l,i}^{(k)}$ and $g_{l,i}^{(k)}$ are the $i$-th elements of
  $\vb{x}_l^{(k)}$ and $\vb{g}_l^{(k)}$, and $\vb{w}_{l,i}$ is the $i$-th
  column of $\vb{W}_l$.
  Citations indicate methods using these measures or close variants thereof.}
  \label{table:sensitivities}
  \renewcommand{\arraystretch}{1.35}
  \begin{tabular}{@{}l@{\hspace{1.5em}}l@{\hspace{1.5em}}l@{}}
    \midrule
    \multicolumn{3}{@{}l}{\textbf{Squared input activation}
      \citep{hu2016network, berisha2025variance}} \\
    \midrule
    \hspace{1em}$\vb{A}_l^{(k)}$: $\vb{I}$
      & $I_l$: $\sum_k \vb{x}_l^{(k)\top} \vb{x}_l^{(k)}$
      & $I_{l,i}$: $\sum_k \big(x_{l,i}^{(k)}\big)^2$ \\
    \addlinespace[8pt]
    \midrule
    \multicolumn{3}{@{}l}{\textbf{Squared (layer) output activation}
      \citep{he2017channel, luo2017thinet, sun2023simple}} \\
    \midrule
    \hspace{1em}$\vb{A}_l^{(k)}$: $\vb{W}_l$
      & $I_l$: $\sum_k \vb{x}_l^{(k)\top} \vb{W}_l^\top \vb{W}_l \vb{x}_l^{(k)}$
      & $I_{l,i}$: $\|\vb{w}_{l,i}\|^2 \sum_k \big(x_{l,i}^{(k)}\big)^2$ \\
    \addlinespace[8pt]
    \midrule
    \multicolumn{3}{@{}l}{\textbf{First-order Taylor}
      \citep{molchanov2016pruning, theis2018faster, molchanov2019importance}} \\
    \midrule
    \hspace{1em}$\vb{A}_l^{(k)}$: $\vb{g}_l^{(k)\top}$
      & $I_l$: $\sum_k \big(\vb{g}_l^{(k)\top} \vb{x}_l^{(k)}\big)^2$
      & $I_{l,i}$: $\sum_k \big(g_{l,i}^{(k)} x_{l,i}^{(k)}\big)^2$ \\
    \addlinespace[8pt]
    \midrule
   \multicolumn{3}{@{}l}{\textbf{Second-order Taylor / Fisher}} \\
   \multicolumn{3}{@{}l}{\citep{lecun1989optimal, hassibi1992second, wang2019eigendamage,
      yang2023global, van2024llm}} \\
      \midrule
    \hspace{1em}$\vb{A}_l^{(k)}$: $\big(\vb{H}_l^{(k)}/2\big)^{1/2} \vb{W}_l$
      & $I_l$: $\tfrac12 \sum_k \vb{x}_l^{(k)\top} \vb{W}_l^\top
         \vb{H}_l^{(k)} \vb{W}_l \vb{x}_l^{(k)}$
      & $I_{l,i}$: $\tfrac12 \sum_k \big(x_{l,i}^{(k)}\big)^2 \,
         \vb{w}_{l,i}^\top \vb{H}_l^{(k)} \vb{w}_{l,i}$ \\
  \end{tabular}
\end{table}

Throughout, we use a single sensitivity shared across samples,
$\vb{S}_l = \frac{1}{K}\sum_{k} \vb{A}_l^{(k)\top}\vb{A}_l^{(k)}$,
so that the layer importance, in matrix form, is
\begin{equation}
  I_l = \operatorname{tr}\big(\vb{X}_l^\top \vb{S}_l \vb{X}_l\big)
      = \operatorname{tr}\big(\vb{S}_l \vb{C}_l\big),
  \qquad \vb{C}_l = \vb{X}_l \vb{X}_l^\top,
\end{equation}
with per-unit importances given by the diagonal terms,
$I_{l,i} = [\vb{S}_l]_{ii}[\vb{C}_l]_{ii}$.
For the activation-based measures and K-FAC-based curvature estimation method the sensitivity does not depend on the sample and this form is exact.
For the other loss-based measures, activities and sensitivities are
averaged separately, which yields the K-FAC form
$\vb{S}_l = \tfrac12\vb{W}_l^\top\bar{\vb{H}}_l\vb{W}_l$, with
$\bar{\vb{H}}_l$ the sample average of the chosen curvature
estimate~\citep{martens2015optimizing, wang2019eigendamage}; for the
first-order measure, this is the empirical Fisher up to a factor of $\flatfrac12$.

A typical pruning pipeline removes the $m$ units of
lowest importance score, together with their outgoing weights.
However, these scores are typically then used per node, individually, as if each node's contribution to this total layer importance were independent and as if $\vb{C}_l$ were diagonal.
In practice, units are often highly redundant in their activity, such that when a unit is removed, much of its contribution could be carried by the units that remain, given appropriate adjustment of their outgoing weights.
In the following, we instead transform the layer so that shared importance is removed from the units to be pruned before any pruning takes place.

\subsection{Removing redundant shared importances}

Prior to pruning, we propose to transform the unit activities of a layer into a latent
basis, $\hat{\vb{X}}_l = \vb{M}_l \vb{X}_l$, with an invertible matrix
$\vb{M}_l$.
Defining the transformed weights and sensitivities
\begin{equation*}
  \tilde{\vb{W}}_l = \vb{W}_l \vb{M}_l^{-1}, \qquad
  \tilde{\vb{A}}_l^{(k)} = \vb{A}_l^{(k)} \vb{M}_l^{-1},
\end{equation*}
we have $\tilde{\vb{W}}_l \hat{\vb{X}}_l = \vb{W}_l \vb{X}_l$ and
$\tilde{\vb{A}}_l^{(k)} \hat{\vb{x}}_l^{(k)} = \vb{A}_l^{(k)} \vb{x}_l^{(k)}$,
so that both the network output and the layer importance $I_l$ are unchanged for any choice of $\vb{A}_l^{(k)}$.

In the latent basis, the layer importance reads
\begin{equation}
  I_l = \operatorname{tr}\big(\tilde{\vb{S}}_l \vb{M}_l \vb{C}_l \vb{M}_l^\top\big)
      = \sum_j \hat{I}_{l,j}
      + \sum_{j \neq j'} \big[\tilde{\vb{S}}_l\big]_{jj'}
        \big[\vb{M}_l \vb{C}_l \vb{M}_l^\top\big]_{jj'},
  \label{eq:latent_decomposition}
\end{equation}
where $\tilde{\vb{S}}_l = \vb{M}_l^{-\top}\vb{S}_l\vb{M}_l^{-1}$, and
$\hat{I}_{l,j} = [\tilde{\vb{S}}_l]_{jj}[\vb{M}_l \vb{C}_l \vb{M}_l^\top]_{jj}$
takes the same form as the per-unit importances $I_{l,i}$.
If the cross terms vanish, the latent importances sum exactly to $I_l$,
each equals the change in $I_l$ caused by removing that latent unit, and
the impact of removing any set of latent units is the sum of their
importances.
\textit{Our goal is therefore to choose $\vb{M}_l$ such that all cross terms vanish and that each latent unit independently contributes to the total layer importance.}

\begin{figure}
    \centering
    \includegraphics[width=0.7\linewidth]{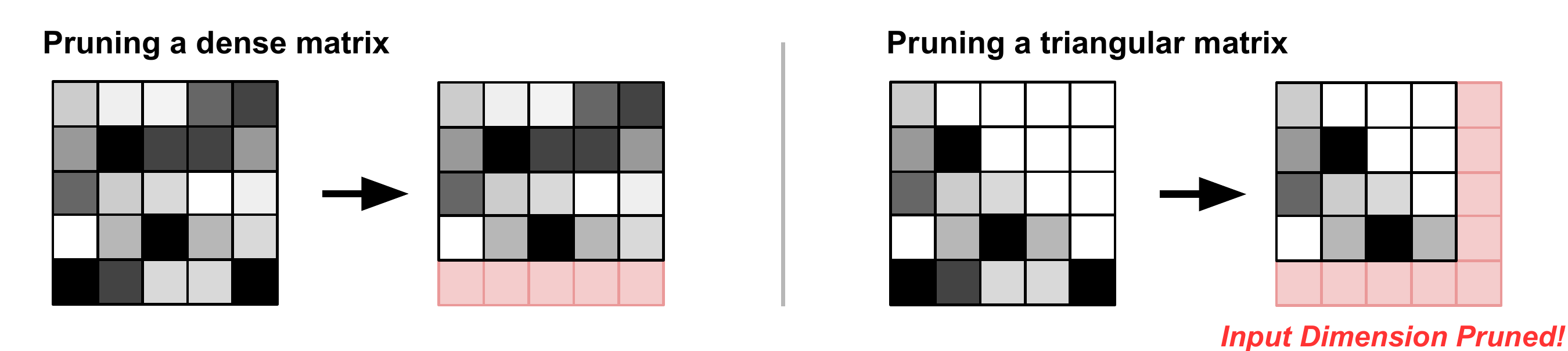}
    \caption{Why the subspace must be constructed with a lower-triangular matrix. Left: with a dense
    $\vb{M}_l$, removing latent units only lowers the rank of the layer (number of nodes in the latent space), and every original input unit remains in use. Right: with a lower-triangular $\vb{M}_l$, removing the bottom row of $\vb{M}_l$ removes the last
    original unit.}
    \label{fig:triangularprune}
\end{figure}

Many transformations could remove these cross terms, for example
orthogonalizing transforms such as PCA or ZCA.
However, pruning in such bases gives a low-rank weight matrix rather
than removing any units~\citep{levin1993fast}.
To remove units, the latent units that are kept must depend only on the
original units that are kept.
Requiring this for pruning the last $n_l - m$ latent units, for every
$m$, restricts $\vb{M}_l$ to be lower-triangular.
Then, with $\mathcal{K} = (1, \dots, m)$ the kept units,
$\hat{\vb{X}}_{l,\mathcal{K}} = [\vb{M}_l]_{\mathcal{K}\mathcal{K}}\vb{X}_{l,\mathcal{K}}$,
and pruning the last $n_l - m$ latent units removes the corresponding
original units (Figure~\ref{fig:triangularprune}).
Note, this setup assumes that our nodes are somehow already ordered such that the last nodes in our original space are the ones we wish to prune first.
We address how to choose this ordering in Section~\ref{sec:pruning-order}.

So, can we find a lower-triangular $\vb{M}_l$ that removes the cross terms?
By Equation~\ref{eq:latent_decomposition}, they vanish whenever
$\vb{M}_l \vb{C}_l \vb{M}_l^\top = \vb{D}_l$ is diagonal, regardless of
$\vb{S}_l$.
For lower-triangular $\vb{M}_l$, which we take to have a unit diagonal
without loss of generality (rescaling a latent unit changes neither the
output nor its importance), rearranging
$\vb{M}_l \vb{C}_l \vb{M}_l^\top = \vb{D}_l$ gives
$\vb{C}_l = \vb{M}_l^{-1} \vb{D}_l \vb{M}_l^{-\top}$.
This is the LDL decomposition of $\vb{C}_l$ with unit lower-triangular
factor $\vb{M}_l^{-1}$, which is unique for a given ordering of the units
when $\vb{C}_l$ is positive definite.
Equivalently, $\vb{M}_l$ performs a Gram--Schmidt orthogonalization of
the unit activities: latent unit $j$ is the residual of original unit
$j$ after regressing out units $1, \dots, j-1$.
The last latent units, which are pruned first, thus retain only activity
that cannot be explained by earlier units, and the diagonal of the
matrix above gives their importances,
$\hat{I}_{l,j} = [\vb{D}_l]_{jj} [\tilde{\vb{S}}_l]_{jj}$.

Pruning then proceeds from the last latent unit backwards, i.e.\ by
removing rows of $\vb{M}_l$ from the bottom.
With $\mathcal{K} = (1, \dots, m)$ the kept and
$\mathcal{P} = (m+1, \dots, n_l)$ the pruned units, the kept latent
activities are
$\hat{\vb{X}}_{l,\mathcal{K}} = [\vb{M}_l]_{\mathcal{K}\mathcal{K}}\vb{X}_{l,\mathcal{K}}$,
so that the pruned layer acts directly on the original kept units, with
weights
\begin{equation}
  \hat{\vb{W}}_l
  = [\tilde{\vb{W}}_l]_{:\mathcal{K}} [\vb{M}_l]_{\mathcal{K}\mathcal{K}}
  = [\vb{W}_l]_{:\mathcal{K}}
  + [\vb{W}_l]_{:\mathcal{P}}\,
    [\vb{C}_l]_{\mathcal{P}\mathcal{K}} \left([\vb{C}_l]_{\mathcal{K}\mathcal{K}}\right)^{-1}.
  \label{eq:pruned_weights}
\end{equation}
The second term is the least-squares reconstruction of the pruned units'
activities from the kept ones, as used for pruning
by~\citet{mariet2015diversity} and~\citet{he2017channel}.
Here, however, it is not a separate step: as the first equality shows, it
follows directly from truncating the lower-triangular basis, such that
pruning and reconstruction become a single operation.
The corresponding rows of $\vb{W}_{l-1}$ are removed.
The kept units take over all contributions of the pruned units that they
can linearly explain, and only the residual activity is discarded.
Since the cross terms vanish, the layer importance decreases by exactly
$\sum_{j \in \mathcal{P}} \hat{I}_{l,j}$, such that the importance lost
after reconstruction is exactly what was known before pruning.
The latent importances computed before pruning are thus exact predictions of the importance lost after reconstruction, unlike standard per-unit scores, which ignore both the cross terms and the compensation by the kept units.
The kept latent units thus span the subspace of the original activities
from which our method takes its name: for the kept units, it retains the
largest possible share of the layer importance, whatever the importance
measure.
In practice, $\vb{C}_l$ may be singular, for example due to inactive or
duplicated units, and one could therefore add a small damping $\lambda\vb{I}$
before the decomposition.

\subsection{Choosing a pruning order}
\label{sec:pruning-order}
The method above assumes that units are ordered from most to least
important, such that those to be pruned first come last in the layer.
To minimize the importance lost by pruning, we seek the permutation of
the units for which the pruned latent units carry the least total
importance, $\min \sum_{j\in\mathcal{P}}\hat{I}_{l,j}$.
Finding this ordering is a combinatorial problem.
We avoid it by scoring each unit once, before the factorization, with a score that does not depend on the order of the other units.
When pruning in several shots, the scores are recomputed at the start of each shot.

Ideally, each unit would be scored by the importance which cannot be
carried by the other units.
Placing a unit last, i.e. at position $n_l$, its latent activity is the residual after
regressing out all other units, with $[\vb{D}_l]_{n_l n_l} = 1/[\vb{C}_l^{-1}]_{n_l n_l}$,
and since the last column of $\vb{M}_l^{-1}$ is the unit vector,
$[\tilde{\vb{S}}_l]_{n_l n_l} = [\vb{S}_l]_{ii}$.
We therefore score each unit by its latent importance as if it had been placed last,
\begin{equation}
  I^{\perp}_{l,i} = \frac{[\vb{S}_l]_{ii}}{[\vb{C}_l^{-1}]_{ii}},
  \label{eq:perp_importance}
\end{equation}
and order the units by decreasing $I^{\perp}_{l,i}$.
Note that this heuristic allows us to avoid the combinatorial problem of checking all possible orders of nodes.

Compared with $I_{l,i} = [\vb{S}_l]_{ii}[\vb{C}_l]_{ii}$, the activity
$[\vb{C}_l]_{ii}$ is replaced by $1/[\vb{C}_l^{-1}]_{ii} \le [\vb{C}_l]_{ii}$,
with equality only if unit $i$ is uncorrelated with all others, so that
the importance of a node which is also shared with other nodes is discounted.
For the K-FAC measure, Equation~\ref{eq:perp_importance} is the Optimal Brain
Surgeon cost of removing an input unit with optimal compensation, as used
in LLM Surgeon~\citep{hassibi1992second, van2024llm}.
The LDL decomposition in the resulting order then removes all remaining
cross terms, as described above.

\subsection{Pruning full networks}
\label{sec:full-networks}
The importances above determine which units to prune within a layer,
but not how to distribute pruning across the layers of a network.
For this, we use that the factorization provides the latent importances
$\hat{I}_{l,j} = [\vb{D}_l]_{jj}[\tilde{\vb{S}}_l]_{jj}$ without extra
computation, and that, since the cross terms vanish, pruning a set of
trailing latent units removes exactly the sum of their importances.
How these are compared across layers depends on the importance measure.

For the activation-based measures, importances are not comparable
across layers, and we instead compare fractions of each layer's
importance.
Pruning the latent units $i, \dots, n_l$ removes the fraction $ \rho_{l,i} = \frac{1}{I_l}\sum_{j=i}^{n_l} \hat{I}_{l,j}$ 
of the layer importance.
We set a single global parameter $\tau$, the fraction of importance to
be removed from every layer, and in each layer prune the largest set of
trailing units with $\rho_{l,i} \le \tau$.
For the squared input activation ($\vb{S}_l = \vb{I}$), $\hat{I}_{l,j}$
reduces to the latent variance $[\vb{D}_l]_{jj}$, and $\rho_{l,i}$ to the
fraction of activity variance removed.

For the loss-based measures, the latent importances estimate the
increase in loss caused by pruning, and are directly comparable across
layers.
We therefore prune globally, as in \citet{van2024llm}: at each step, among
the trailing latent units of all layers, we prune the one with the
smallest importance, until the target FLOPs reduction is reached.
Within a layer, the importance lost is exactly the sum of the pruned
latent importances, while across layers this sum neglects interactions
between layers.
The full pipeline is given in Algorithm~\ref{alg:layerwise_pruning}.

\begin{algorithm}[t]
\caption{Pruning a network in the latent basis.}
\label{alg:layerwise_pruning}
\begin{small}
\begin{algorithmic}[1]
\Require weights $\{\vb{W}_l\}$, calibration data, damping $\lambda$,
  number of shots $T$, fraction $\tau$ or FLOPs target $F$
\For{shot $t = 1, \dots, T$}
  \State Compute $\{\vb{X}_l\}$, $\{\vb{S}_l\}$ on the current network
  \For{each layer $l$}
    \State $\vb{C}_l \gets \vb{X}_l\vb{X}_l^\top$ \Comment{if necessary adding $\lambda \vb{I}$}
    \State Reorder units by decreasing $I^{\perp}_{l,i}$
      \Comment{Equation~\ref{eq:perp_importance}}
    \State $\vb{L}_l\vb{D}_l\vb{L}_l^\top \gets \vb{C}_l$,
      \; $\hat{I}_{l,j} \gets [\vb{D}_l]_{jj}[\vb{L}_l^\top\vb{S}_l\vb{L}_l]_{jj}$
      \Comment{LDL, $\vb{L}_l = \vb{M}_l^{-1}$}
  \EndFor
  \State Select trailing units $\mathcal{P}_l$ per layer using $\tau$
    or a shot specific target (interpolated $F$) \Comment{Section~\ref{sec:full-networks}}
  \For{each layer $l$}
    \State Update $\vb{W}_l$ by Equation~\ref{eq:pruned_weights};
      remove rows $\mathcal{P}_l$ of $\vb{W}_{l-1}$
  \EndFor
\EndFor
\end{algorithmic}
\end{small}
\end{algorithm}

\section{Experimental Setup}
\paragraph{Networks and datasets}
To demonstrate the efficacy of our proposed method, we first apply it to
the single-branch models VGG-16~\citep{simonyan2014very},
followed by ResNet-50~\citep{he2016deep}, and lastly the transformer
architectures OPT~\citep{zhang2022opt}, and LLaMA-2 7B~\citep{touvron2023llama}.
OPT and LLaMA-2 are evaluated on the WikiText
dataset~\citep{merity2016pointer}.
All other models are pretrained and evaluated on the ILSVRC 2012
(ImageNet) dataset~\citep{russakovsky2015imagenet}.
The pruning procedure on LLMs was based entirely upon the openly available codebase of LLM Surgeon with matched hyperparameters, samples, sequence lengths, and sparsity target computations.
Modified code thereof and all code used to produce the results herein is available at \href{https://github.com/EfficientNeuralSystems/subspace-node-pruning}{https://github.com/EfficientNeuralSystems/subspace-node-pruning}.

\paragraph{Prunable units}
For the vision models, we apply Algorithm~\ref{alg:layerwise_pruning} to every prunable layer, with the input units of each linear transform as the units to be pruned, as described in Section~\ref{sec:factorising}.
For convolutional layers, the input is unfolded into patches, and an input channel then corresponds to a group of $k^2$ units, which are kept or pruned together.
For the language models, we follow the structured pruning protocol of LLM Surgeon~\citep{van2024llm}, which prunes both the rows (outputs) and the columns (inputs) of every linear transform in the attention and MLP blocks, with token positions treated as samples.
As in~\citet{van2024llm}, for dimensions shared with the residual stream and across query, key, and value projections, pruned rows and columns are set to zero rather than removed and we report sparsity as the fraction of parameters set to zero.

\paragraph{Metrics and hardware}
We measured performance as the Top-1 test accuracy for vision models and
perplexity for language models, in relation to their FLOPs, and parameter
count.
We used the fvcore package
(\url{https://github.com/facebookresearch/fvcore}) to estimate the FLOPs.
All experiments were run on between one and four Nvidia A100 GPU machines.

\section{Results}
\label{sec:results}

\subsection{Pruning order}
\label{sec:res-order}

\begin{figure}[!ht]
    \centering
    \begin{minipage}[b]{0.68\linewidth}
        \centering
        \includegraphics[height=3.5cm, width=\linewidth, keepaspectratio, trim=0 10 0 0, clip]{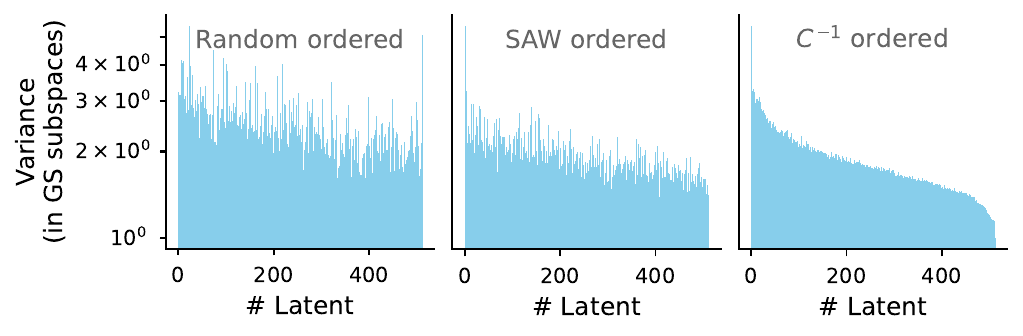}\\
        \small (a) Effect of node ordering
    \end{minipage}\hfill
    \begin{minipage}[b]{0.32\linewidth}
        \centering
        \includegraphics[height=3.5cm, width=\linewidth, keepaspectratio]{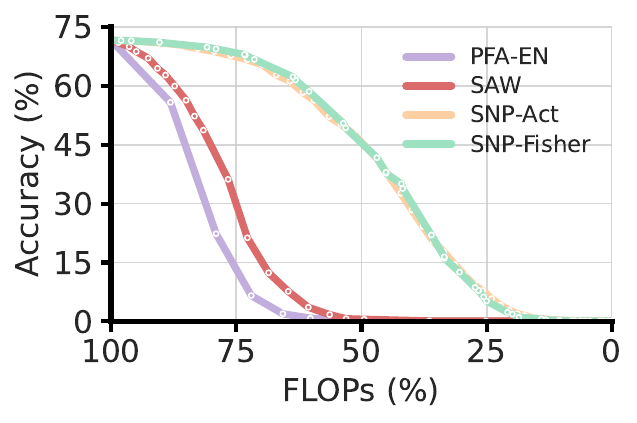}\\
        \small (b) Accuracy after pruning
    \end{minipage}
    \caption{\textbf{Node ordering and pruning performance on VGG-16.}
    \textbf{(a)} Subspace importances $\hat{I}_{l,j}$ of layer 12,
    for units ordered randomly (left), by the summed absolute weights
    (SAW) of~\citet{li2016pruning} (middle), and by our score
    $I^{\perp}_{l,i}$ (right). Units to the
    right are pruned first.
    \textbf{(b)} Accuracy over a range of pruning ratios for our method
    (SNP) and baseline methods (Appendix~\ref{app:hyp-overview}).
    SNP-Act uses the squared input activation with the $I^{\perp}$ ordering, and SNP-Fisher is SNP with the empirical Fisher. PFA-EN is the only baseline
    that uses PCA to determine global importance.}
    \label{fig:vgg-combined}
\end{figure}

Because our latent importances are additive, the importance lost by pruning
is the sum over the trailing latent units, so a good ordering places the
least important latent units last.
Figure~\ref{fig:vgg-combined}a shows that the ordering
determines how well this is achieved.
Randomly ordering units before transformation to our proposed subspace gives little relationship between the ordering of the latent bases and their importance measure.
Ordering units by the summed absolute weight (SAW) importance heuristic~\citep{li2016pruning} before the subspace transformation improves on using a random order, but large latent importances are still distributed across the ordering.
Ordering by $I^{\perp}_{l,i}$ (Equation~\ref{eq:perp_importance}) instead gives
latent importances that more smoothly reduce from first to last unit, such that the units pruned first carry the least importance.
Although the ordering is obtained from a single scoring rather than a combinatorial search (Section~\ref{sec:pruning-order}), it nonetheless closely follows the redundancy captured by our importance measure and works well in practice.

\subsection{Convolutional network prunings}
\label{sec:res-conv}

\paragraph{Single-branch networks}
Figure~\ref{fig:vgg-combined}b compares our method (SNP with activity vs Fisher derived importance scores) against baseline methods on
VGG-16 without any finetuning (see
Appendix~\ref{app:hyp-overview} for baseline details).
SNP retains most of the original accuracy, degrading only slowly as pruning increases. At 50\% of FLOPs pruned, competing non-reconstructive methods have reached close to chance, where SNP methods maintain accuracies above 40\%.

\begin{wrapfigure}{r}{0.4\textwidth}
    \centering
    \vspace{-10pt}
    \includegraphics[width=0.35\textwidth]{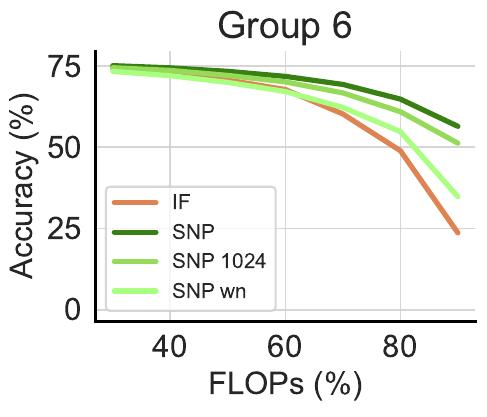}
    \caption{\textbf{SNP vs.\ Intra-Fusion for reconstruction.}
    A comparison of pruning different groups of layers of ResNet-50.
    We compare our SNP method with the squared input activation measure against
    Intra-Fusion (IF). SNP 1024 and SNP wn estimate $\vb{C}_l$ from 1024
    images and from white noise inputs, respectively, to show robustness to the samples used.
    SNP's LLS reconstruction outperforms IntraFusions Optimal Transport approach over the range tested.}
    \label{fig:intra-fusion}
    \vspace{-25pt}
\end{wrapfigure}

\paragraph{Multi-branch networks}
Figure~\ref{fig:intra-fusion} compares SNP with Intra-Fusion (IF)~\citep{theus2024towards}, a method which uses optimal transport as a mechanism for reconstruction, across pruning groups of ResNet-50.
SNP outperforms IF for all groups, and continues to do so when $\vb{C}_l$ is estimated from only 1024 images or even white noise inputs.
This demonstrates the efficacy of the linear least squares reconstruction embedded within our subspace construction.

\subsection{Language models}
\label{sec:res-llm}

\begin{table}
  \centering
  \small
  \caption{\textbf{Pruning LLMs without retraining.} Perplexity on WikiText-2 after structured pruning of rows and columns with SNP and LLM Surgeon~\citep{van2024llm}, both with the empirical-Fisher K-FAC curvature, for different numbers of shots and sparsities (fraction of parameters removed). Lower is better; the better method in each pair is in bold. Results on OPT-125M for all numbers of shots are given in Table~\ref{table:llm-full}.}
  \label{table:llm}
  \begin{tabular}{@{}cc|rr|rr@{}}
    \toprule
    & & \multicolumn{2}{c|}{\textbf{OPT-125M}} & \multicolumn{2}{c}{\textbf{LLaMA-2 7B}} \\
    \textbf{Shots} & \textbf{Sparsity} & SNP & LLM Surgeon & SNP & LLM Surgeon \\
    \midrule
    \multicolumn{2}{c|}{Dense} & \multicolumn{2}{c|}{27.65} & \multicolumn{2}{c}{5.12} \\
    \midrule
    1 & 10\% & \textbf{28.48} & 29.30 & \textbf{5.89} & 7237 \\
     & 30\% & \textbf{55.13} & 3357 & \textbf{19.14} & 4791 \\
     & 50\% & \textbf{579.72} & 4258 & \textbf{241.77} & 13169 \\
    \midrule
    5 & 10\% & \textbf{27.85} & 28.97 & 5.49 & \textbf{5.27} \\
     & 30\% & \textbf{32.41} & 33.54 & \textbf{8.27} & 12.13 \\
     & 50\% & \textbf{54.21} & 59.79 & \textbf{19.00} & 7986 \\
    \midrule
    40 & 50\% & \textbf{48.89} & 49.72 & \textbf{14.88} & 15.67 \\
    \bottomrule
  \end{tabular}
\end{table}

Table~\ref{table:llm} compares SNP against LLM Surgeon~\citep{van2024llm} on
OPT-125M and LLaMA-2 7B.
LLM Surgeon is run with its authors' own codebase, with the same
calibration data as SNP.
To verify this setup, we compare our 40-shot runs of LLM Surgeon with the values reported by~\citet{van2024llm} in Appendix~\ref{app:llm-baselines} and show that these match.
The degradation of LLM Surgeon at low shot-number is therefore a previously underdiscussed failure mode.
Both methods use the same empirical-Fisher curvature estimation, unit ordering, and weight compensation, and prune over a number of shots, re-estimating the curvature and the activity correlations (importances) after each.
They differ only in how pruning costs combine: LLM Surgeon selects units by their individual costs, whereas SNP selects them by latent importances that are exactly additive within a layer (Section~\ref{sec:full-networks}).

Across all cases bar one, regardless of the number of shots allowed, SNP leads in maintaining lower perplexity.
Over an even more extended set of 60 settings tested on OPT-125M SNP achieves the lower perplexity in 59, see Appendix~\ref{app:opttable}, Table~\ref{table:llm-full}.
LLM Surgeon's per-unit costs neglect the correlations among units that are pruned together, an error that grows with the number of units pruned together.
Thus, LLM Surgeon's error is mitigated with more shots explaining why its performance is robust only in the large-shot regime.
Since the latent importances of SNP already account for activity correlations, SNP needs far fewer shots for a given perplexity, and each shot saved avoids a full re-estimation of the curvature.

\begin{figure}
    \centering
    \includegraphics[width=1.0\linewidth]{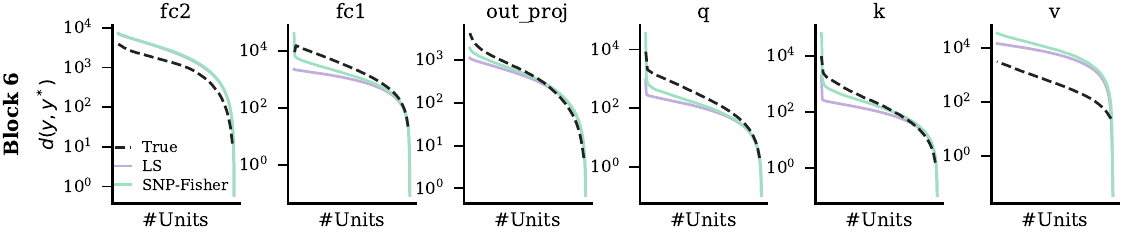}
    \caption{\textbf{True Loss Changes vs Loss Predictions of SNP vs LLM Surgeon.} These plots show the loss change induced by pruning units (more units pruned from right to left) using the SNP with empirical Fisher vs LLM Surgeon (LS) method. These loss magnitude changes are shown for the middle block of an OPT-125M model and for all of its linear layers including two fully connected (fc) layers of the MLP, an output projection layer (out\_proj), as well as the Q, K, and V matrices.}
    \label{fig:opt_linear_main}
\end{figure}

To further illuminate the differences between SNP's (with empirical Fisher) and LLM Surgeon's importance measures, we plot these methods' predicted vs actual loss change for single shot pruning in the OPT-125M model.
Figure~\ref{fig:opt_linear_main} shows the impact of pruning inputs of the middle block of the OPT-125M model's fully connected (fc), output projecting (out\_proj) and QKV matrices.
As can be seen, SNP's predictions of loss difference for each node are equivalent to LLM Surgeon for the trailing units (where their measures are equivalent) but as one prunes more units (left to rigth), the impact of one node's pruning on other nodes increases and shows a deviation of our method vs LLM Surgeon.
In five of the six plotted matrices, our deviation matches more closely to the true loss cost of pruning where LLM Surgeon underestimates it at high pruning ratios.
More extended analysis of this is provided in Appendix~\ref{app:optcurves}.
Appendix~\ref{app:llm-baselines} also provides further evidence of SNP's shot-level benefits against alternative LLM pruning methods such as SliceGPT~\citep{ashkboos2024slicegpt}, K-OBD~\citep{lecun1989optimal, wang2019eigendamage}, and LLM Surgeon.


\section{Discussion}
Our framework builds on and unifies several existing lines of work.
\citet{mariet2015diversity} and \citet{he2017channel} reconstructed
pruned units by least squares as a separate step; in our subspace, this
reconstruction follows directly from pruning within the subspace.
K-FAC-based methods such as LLM Surgeon~\citep{van2024llm} provide
effective loss-based importances, which our framework adopts as one of many potential
choices of sensitivity.
However, where other methods score units individually, our subspace
importances are exactly additive within a layer, such that the joint
cost of pruning many correlated units is known.
Our framework is also closely related to greedy sequential pruning with optimal compensation~\citep{frantar2022optimal, kurtic2023ziplm}.
For a fixed order of removal, these sequential costs are precisely our latent importances and yet our importances are cheaper to compute.

Nonetheless, several questions remain open.
First, the additivity of importances holds only within a layer.
Pruning across layers introduces interactions between layers that our
importances do not capture.
Accounting for these interactions, for example through block-level
curvature estimates, is a natural next step as well as giving block-level importances more focus.
Second, we order units by a single-pass score, $I^{\perp}_{l,i}$ (Equation~\ref{eq:perp_importance}), rather than solving the combinatorial problem of finding the optimal ordering.
Greedy sequential methods~\citep{frantar2022optimal, kurtic2023ziplm} choose the order adaptively instead, at the cost of an update per removed unit.
Better single-step approximations of this ordering could aim to bridge the gap in computational cost and accuracy between these.

Overall, our results suggest that much of the structured-pruning error comes from treating the importances of correlated units as independent.
Measuring importance in a basis where it is additive removes this error, reduces the number of costly curvature re-estimations needed to prune large models, and remains effective even when the correlations are estimated from small data samples.
As importance measures continue to improve, we expect that computing them in such a subspace is a cheap and broadly applicable way to make them reliable when many units are pruned at once.






\bibliography{references}
\bibliographystyle{iclr2027_conference}

\newpage
\appendix
\section*{Appendix}

\section{Experimental details}
\label{app:hyp-overview}

\paragraph{Models}
For our experiments on VGG-16 and ResNet-50, we prune the pretrained networks from PyTorch~\citep{paszke2019pytorch}, trained on the ILSVRC 2012 (ImageNet) dataset using \texttt{IMAGENET1K\_V1}.
The pretrained OPT-125M and LLaMA-2 7B models were loaded from Hugging
Face~\citep{wolf2019huggingface}.

\paragraph{VGG-16 baselines}
Several comparison methods are re-implemented, including summed absolute
weights (SAW)~\citep{li2016pruning} and
PFA-EN~\citep{cuadros2020filter}.
SAW does not provide guidance on the global ranking of units,
instead assuming that a practitioner might prune all layers uniformly by
the same amount.
We therefore use a uniform pruning ratio across layers when implementing
these methods.
In contrast, PFA-EN performs PCA to decide on a global ranking of units
on top of their local importance structure.
We apply PFA-EN at the input units rather than the output activations,
as we found this to give the best performance.
We re-implemented these baselines due to unavailable code or outdated
packages.

\paragraph{Hyperparameters}
Table~\ref{tab:hyperparameters} lists the hyperparameters for all
experiments. No fine-tuning is performed after pruning.
For the vision models, all pruning is done in a single shot ($T = 1$).
For the language models, we use the calibration data and target sizes of
LLM Surgeon~\citep{van2024llm}, where the target size $\alpha$ is the
fraction of parameters that remains after pruning.

\begin{table}[h]
  \centering
  \small
  \caption{Hyperparameters for all experiments. For the language models,
  the damping follows the LLM Surgeon defaults~\citep{van2024llm}, with
  $\lambda_C$ and $\lambda_H$ added to $\vb{C}_l$ and $\bar{\vb{H}}_l$
  relative to their diagonals.}
  \label{tab:hyperparameters}
  \begin{tabular}{@{}lllll@{}}
    \toprule
    & VGG-16 & ResNet-50 & OPT-125M & LLaMA-2 7B \\
    \midrule
    Pretrained weights
      & torchvision & torchvision
      & Hugging Face & Hugging Face \\
    Calibration data
      & ImageNet train & ImageNet train
      & WikiText-2 train & WikiText-2 train \\
    Calibration samples
      & Images & Images
      & 128 sequences & 128 sequences \\
    Sequence length
      & -- & -- & 2048 tokens & 2048 tokens \\
    Evaluation data
      & ImageNet val & ImageNet val
      & WikiText-2 test & WikiText-2 test \\
    Shots $T$
      & 1 & 1 & up to 40 & up to 40 \\
    Pruning target
      & - & -
      & $\alpha \in \{0.9, \dots, 0.5\}$
      & $\alpha \in \{0.9, \dots, 0.5\}$ \\
    Damping $\lambda_C$, $\lambda_H$
      & 0, -- & 0, --
      & 0.01, 0.01 & 0.01, 0.01 \\
    Fine-tuning
      & none & none & none & none \\
    \bottomrule
  \end{tabular}
\end{table}
\newpage
\newpage

\section{Extended OPT-125M Results}
\label{app:opttable}
\begin{table}[!ht]
  \centering
  \footnotesize
  \setlength{\tabcolsep}{3.5pt}
  \caption{Perplexity on WikiText-2 of OPT-125M (dense: 27.65) after structured pruning of rows and columns with SNP and LLM Surgeon (LS), for all numbers of shots and sparsities, extending Table~\ref{table:llm}. Lower is better; the better method in each pair is in bold.}
  \label{table:llm-full}
  \begin{tabular}{@{}c|cc|cc|cc|cc|cc@{}}
    \toprule
    & \multicolumn{2}{c|}{10\%} & \multicolumn{2}{c|}{20\%} & \multicolumn{2}{c|}{30\%} & \multicolumn{2}{c|}{40\%} & \multicolumn{2}{c}{50\%} \\
    \textbf{Shots} & SNP & LS & SNP & LS & SNP & LS & SNP & LS & SNP & LS \\
    \midrule
    1 & \textbf{28.48} & 29.30 & \textbf{34.95} & 221.06 & \textbf{55.13} & 3357 & \textbf{132.01} & 3875 & \textbf{579.72} & 4258 \\
    2 & \textbf{27.93} & 28.70 & \textbf{30.01} & 32.01 & \textbf{34.89} & 44.59 & \textbf{45.13} & 270.25 & \textbf{66.19} & 2699 \\
    3 & \textbf{27.87} & 30.15 & \textbf{29.08} & 31.33 & \textbf{34.14} & 34.97 & \textbf{43.21} & 50.92 & \textbf{63.08} & 106.89 \\
    4 & \textbf{27.98} & 28.31 & \textbf{29.64} & 30.51 & \textbf{34.19} & 37.13 & \textbf{41.40} & 45.07 & \textbf{56.47} & 69.32 \\
    5 & \textbf{27.85} & 28.97 & \textbf{29.07} & 31.50 & \textbf{32.41} & 33.54 & \textbf{41.06} & 45.05 & \textbf{54.21} & 59.79 \\
    10 & \textbf{27.94} & 28.34 & \textbf{28.84} & 30.19 & \textbf{32.37} & 33.57 & \textbf{39.57} & 43.21 & \textbf{52.91} & 55.32 \\
    15 & \textbf{27.86} & 28.22 & \textbf{28.77} & 31.05 & \textbf{32.23} & 33.51 & \textbf{38.68} & 42.09 & \textbf{49.72} & 57.33 \\
    20 & \textbf{27.92} & 28.00 & \textbf{28.80} & 29.41 & \textbf{32.11} & 33.16 & \textbf{38.22} & 39.50 & \textbf{49.98} & 50.41 \\
    25 & \textbf{27.81} & 28.01 & \textbf{28.65} & 30.52 & \textbf{31.82} & 33.67 & \textbf{38.32} & 40.62 & \textbf{49.82} & 51.54 \\
    30 & \textbf{27.91} & 27.99 & \textbf{28.73} & 29.10 & \textbf{31.90} & 32.62 & \textbf{38.32} & 41.09 & \textbf{50.60} & 50.88 \\
    35 & 28.13 & \textbf{28.03} & \textbf{28.61} & 28.81 & \textbf{31.86} & 38.54 & \textbf{37.49} & 37.77 & \textbf{48.53} & 50.17 \\
    40 & \textbf{27.87} & 28.10 & \textbf{28.62} & 28.71 & \textbf{31.74} & 31.88 & \textbf{37.73} & 38.67 & \textbf{48.89} & 49.72 \\
    \bottomrule
  \end{tabular}
\end{table}

\newpage
\section{Comparison with reported LLM baselines}
\label{app:llm-baselines}
Table~\ref{table:llm-baselines} compares SNP with results reported for
SliceGPT~\citep{ashkboos2024slicegpt}, for K-OBD~\citep{lecun1989optimal, wang2019eigendamage}
and LLM Surgeon as reported by~\citet{van2024llm}.
K-OBD uses the same Kronecker-factored curvature as LLM Surgeon, but scores
units independently and does not compensate the remaining weights.
Both are reported from a single run with a linear 40-shot schedule to 50\%
sparsity, such that lower sparsities are reached after proportionally fewer
shots (8 shots for 10\%, 16 for 20\%, and so on).
SliceGPT is a single-shot method that removes embedding dimensions rather
than units, such that its slicing fraction does not correspond to the
fraction of parameters removed; we report its results with 128 calibration
sequences, and its evaluation gives a different dense perplexity for
LLaMA-2 7B.

\begin{table}[!ht]
  \centering
  \footnotesize
  \setlength{\tabcolsep}{4pt}
    \caption{Perplexity on WikiText-2 of SNP with 5 and 40 shots, compared
    with results reported for SliceGPT~\citep{ashkboos2024slicegpt}, and for
    K-OBD and LLM Surgeon (LS) by~\citet{van2024llm}, on OPT-125M (top) and
    LLaMA-2 7B (bottom). Lower is better; the best value in each row is in
    bold, and -- marks settings not evaluated. $^\dagger$Values in italics are
    reported by other works. Numbers in parentheses give the number of shots
    used by each reported method: SliceGPT is single-shot, and K-OBD and LS are
    reported from a single run with a linear 40-shot schedule to 50\% sparsity,
    reaching lower sparsities after fewer shots. Only at 50\% sparsity do the
    reported K-OBD and LS results use the same number of shots as SNP with
    40 shots; all other comparisons are indicative only.}
  \label{table:llm-baselines}

  \begin{tabular}{@{}c|ccccc@{}}
    \toprule
    & \multicolumn{5}{c}{\textbf{OPT-125M}} \\
    & \multicolumn{2}{c}{SNP} & & & \\
    \cmidrule(lr){2-3}
    \textbf{Sparsity} & 5 shots & 40 shots & SliceGPT$^\dagger$ & K-OBD$^\dagger$ & LS$^\dagger$ \\
    \midrule
    Dense & \multicolumn{2}{c}{27.65} & \textit{27.64} & \textit{27.65} & \textit{27.65} \\
    \midrule
    10\% & \textbf{27.85} & 27.87 & \textit{29.33} (1) & \textit{27.97} (8) & \textit{28.01} (8) \\
    20\% & 29.07 & \textbf{28.62} & \textit{34.53} (1) & \textit{29.89} (16) & \textit{28.73} (16) \\
    30\% & 32.41 & \textbf{31.74} & \textit{44.61} (1) & \textit{36.54} (24) & \textit{31.82} (24) \\
    40\% & 41.06 & \textbf{37.73} & -- & \textit{47.54} (32) & \textit{38.47} (32) \\
    50\% & 54.21 & \textbf{48.89} & -- & \textit{75.95} (40) & \textit{49.78} (40) \\
    \bottomrule
  \end{tabular}

  \vspace{1em}

  \begin{tabular}{@{}c|ccccc@{}}
    \toprule
    & \multicolumn{5}{c}{\textbf{LLaMA-2 7B}} \\
    & \multicolumn{2}{c}{SNP} & & & \\
    \cmidrule(lr){2-3}
    \textbf{Sparsity} & 5 shots & 40 shots & SliceGPT$^\dagger$ & K-OBD$^\dagger$ & LS$^\dagger$ \\
    \midrule
    Dense & \multicolumn{2}{c}{5.12} & \textit{5.47} & \textit{5.12} & \textit{5.12} \\
    \midrule
    10\% & 5.49 & -- & \textit{5.96} (1) & \textit{5.48} (8) & \textbf{\textit{5.25}} (8) \\
    30\% & 8.27 & -- & \textit{8.64} (1) & \textit{15.43} (24) & \textbf{\textit{7.83}} (24) \\
    50\% & 19.00 & \textbf{14.88} & -- & \textit{46.64} (40) & \textit{15.38} (40) \\
    \bottomrule
  \end{tabular}
\end{table}

These comparisons differ in how directly they can be interpreted.
At 50\% sparsity, SNP with 40 shots and the reported K-OBD and LLM Surgeon
results use the same number of shots, and our pipeline is built on the
LLM Surgeon codebase with matched settings; our own 40-shot runs of
LLM Surgeon reproduce the reported values closely (Section~\ref{sec:res-llm}).
These results are therefore directly comparable.
At lower sparsities, SNP uses either fewer (5) or more (40) shots than the
reported runs, and the comparison with SliceGPT involves a different pruning
structure and evaluation; these comparisons are indicative only.

In the directly comparable setting, SNP achieves the lowest perplexity on both
models: 48.89 against 49.78 for LLM Surgeon on OPT-125M, and 14.88 against
15.38 on LLaMA-2 7B, with K-OBD far behind on both.
On OPT-125M, SNP with 40 shots also achieves the lowest perplexity at every
lower sparsity, and SNP with only 5 shots outperforms K-OBD and SliceGPT at
all reported sparsities.
On LLaMA-2 7B, LLM Surgeon with 8 and 24 shots achieves lower perplexity than
SNP with 5 shots at 10\% and 30\% sparsity, respectively (5.25 against 5.49,
and 7.83 against 8.27).
With 5 shots, SNP matches K-OBD at 10\% sparsity (5.49 against 5.48),
outperforms it substantially at 30\% (8.27 against 15.43), and outperforms
SliceGPT at both.

\newpage
\subsection{OPT-125M Predicted vs True Loss Impacts of Pruning}
\label{app:optcurves}

\begin{figure}[!ht]
    \centering
    \includegraphics[width=1.0\linewidth]{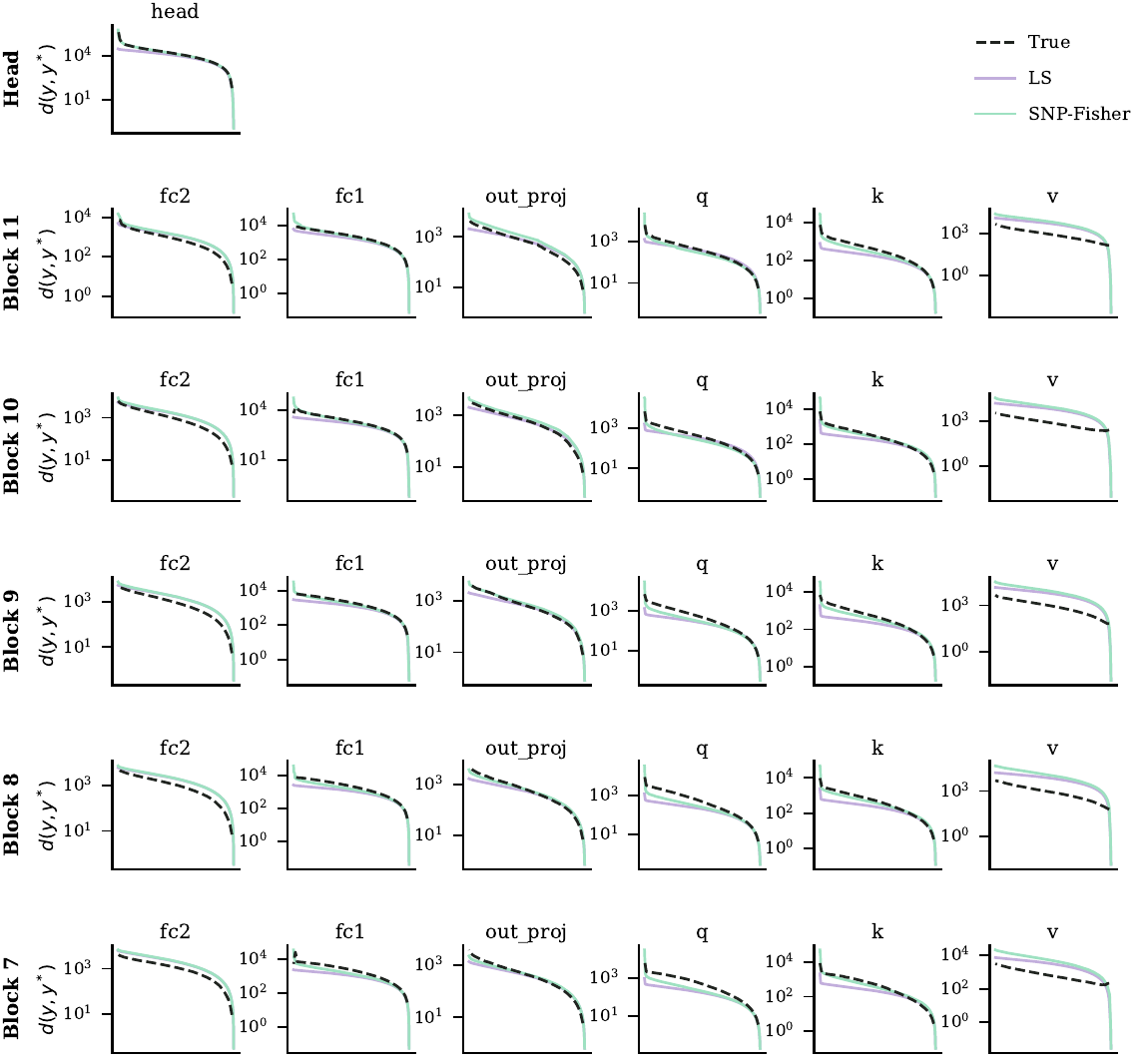}
    \caption{\textbf{Predicted and true loss change when pruning units of
    all blocks of OPT-125M.}
    Rows correspond to blocks 7-12 and columns to the linear layers of
    each block (fc1, fc2, out\_proj, and the Q, K and V projections).
    Units are pruned from right to left in a single shot; curves show the
    true change in loss and the change predicted by SNP-Fisher and by
    LLM Surgeon.
    As in the main text, the SNP-Fisher prediction stays close to
    the true loss as more units are pruned, whereas LLM Surgeon's
    prediction diverges.}
    \label{fig:opt_linear_all_top}
\end{figure}

\begin{figure}[!ht]
    \centering
    \includegraphics[width=1.0\linewidth]{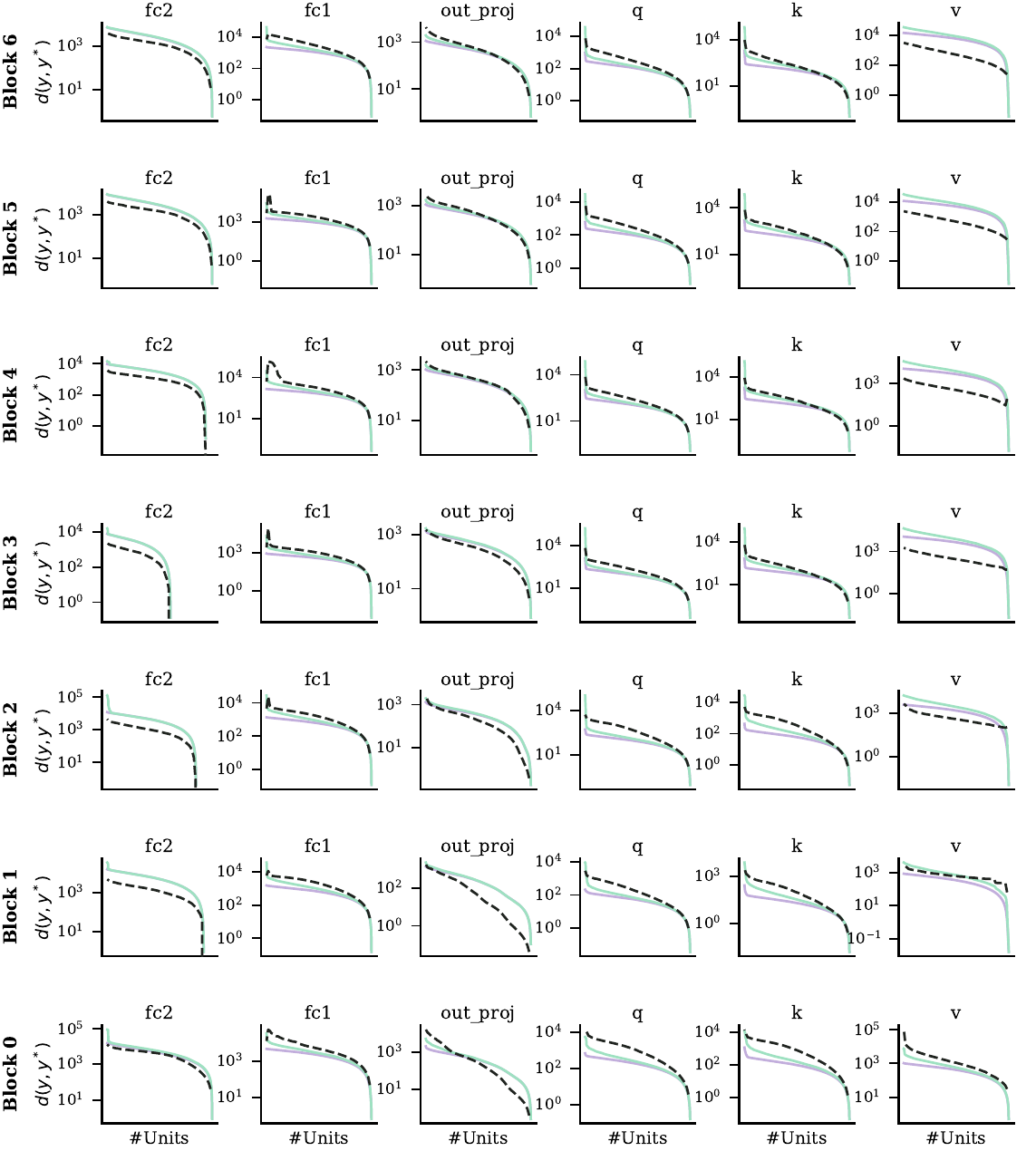}
    \caption{\textbf{Predicted and true loss change when pruning units of
    all blocks of OPT-125M.}
    Rows correspond to blocks 1-6 and columns to the linear layers of
    each block (fc1, fc2, out\_proj, and the Q, K and V projections).
    Units are pruned from right to left in a single shot; curves show the
    true change in loss and the change predicted by SNP-Fisher and by
    LLM Surgeon.
    As in the main text, the SNP-Fisher prediction stays close to
    the true loss as more units are pruned, whereas LLM Surgeon's
    prediction diverges.}
    \label{fig:opt_linear_all_bot}
\end{figure}

Figure~\ref{fig:opt_linear_all_top} and~\ref{fig:opt_linear_all_bot} extends Figure~\ref{fig:opt_linear_main}
to all 12 blocks of OPT-125M. For each linear layer, units are pruned
in a single shot, and we compare the true change in loss with the change
predicted by SNP-Fisher and by LLM Surgeon.

\end{document}

%% file: math_commands.tex
\usepackage{amsmath,amsfonts,bm}

\def\eqref#1{equation~\ref{#1}}

\def\1{\bm{1}}

\def\vb{{\bm{b}}}

\DeclareMathAlphabet{\mathsfit}{\encodingdefault}{\sfdefault}{m}{sl}
\SetMathAlphabet{\mathsfit}{bold}{\encodingdefault}{\sfdefault}{bx}{n}

